\documentclass[letterpaper]{article} % DO NOT CHANGE THIS
\usepackage{aaai24}  % DO NOT CHANGE THIS
\usepackage{times}  % DO NOT CHANGE THIS
\usepackage{helvet}  % DO NOT CHANGE THIS
\usepackage{courier}  % DO NOT CHANGE THIS
\usepackage[hyphens]{url}  % DO NOT CHANGE THIS
\usepackage{graphicx} % DO NOT CHANGE THIS
\usepackage{natbib}  % DO NOT CHANGE THIS AND DO NOT ADD ANY OPTIONS TO IT
\usepackage{caption} % DO NOT CHANGE THIS AND DO NOT ADD ANY OPTIONS TO IT
\usepackage{algorithm}
\usepackage{algorithmic}

\usepackage{newfloat}
\usepackage{listings}
\DeclareCaptionStyle{ruled}{labelfont=normalfont,labelsep=colon,strut=off} % DO NOT CHANGE THIS
\floatstyle{ruled}
\newfloat{listing}{tb}{lst}{}
\floatname{listing}{Listing}

\usepackage{amsmath}
\usepackage{amsfonts}
\usepackage{amssymb}
\usepackage{dsfont}
\usepackage{enumitem}
\usepackage{array}
\usepackage{booktabs}
\usepackage[table]{xcolor}

\title{Towards Transferable Adversarial Attacks with Centralized Perturbation}
\author{
    Shangbo Wu\textsuperscript{\rm 1},
    Yu-an Tan\textsuperscript{\rm 1},
    Yajie Wang\textsuperscript{\rm 1},
    Ruinan Ma\textsuperscript{\rm 1},
    Wencong Ma\textsuperscript{\rm 2},
    Yuanzhang Li\textsuperscript{\rm 2}\thanks{Corresponding author.}
}
\affiliations{
    \textsuperscript{\rm 1}School of Cyberspace Science and Technology, Beijing Institute of Technology\\
    \textsuperscript{\rm 2}School of Computer Science and Technology, Beijing Institute of Technology\\

    \{shangbo.wu, tan2008, wangyajie19, ruinan, wencong.ma, popular\}@bit.edu.cn%
}

\begin{document}

\maketitle

\begin{abstract}
    Adversarial transferability enables black-box attacks on unknown victim deep neural networks (DNNs), rendering attacks viable in real-world scenarios. Current transferable attacks create adversarial perturbation over the entire image, resulting in excessive noise that overfit the source model. Concentrating perturbation to dominant image regions that are model-agnostic is crucial to improving adversarial efficacy. However, limiting perturbation to local regions in the spatial domain proves inadequate in augmenting transferability. To this end, we propose a transferable adversarial attack with fine-grained perturbation optimization in the frequency domain, creating centralized perturbation. We devise a systematic pipeline to dynamically constrain perturbation optimization to dominant frequency coefficients. The constraint is optimized in parallel at each iteration, ensuring the directional alignment of perturbation optimization with model prediction. Our approach allows us to centralize perturbation towards sample-specific important frequency features, which are shared by DNNs, effectively mitigating source model overfitting. Experiments demonstrate that by dynamically centralizing perturbation on dominating frequency coefficients, crafted adversarial examples exhibit stronger transferability, and allowing them to bypass various defenses.
\end{abstract}

\section{Introduction}

DNNs demonstrate outstanding performance across various real-world applications~\cite{HeZRS16,DosovitskiyB0WZ21}. However, DNNs remain susceptible to adversarial examples --- malicious samples with carefully-crafted imperceptible perturbation that disrupt DNN functionalities~\cite{ZhangTSZZL23}. The transferability of adversarial examples~\cite{Liu2016DelvingIT,GuoZLKZHT19} allows for cross-model black-box attacks on even unknown victim DNNs, i.e., perturbation created on one model can fool another with no modification, posing a practical real-world threat.

Existing transferable adversarial attacks are gradient-based iterative attacks, stemming from FGSM~\cite{GoodfellowSS14} and PGD~\cite{MadryMSTV18}. By greedily accumulating gradient information obtained from the white-box source model, these attacks are able to generate transferable adversarial perturbation. However, as attacks try to search the entirety of the input space, the perturbation created tend to overfit the source model, producing excessive noise.

\begin{figure}[t]
    \centering
    \includegraphics[width=\linewidth]{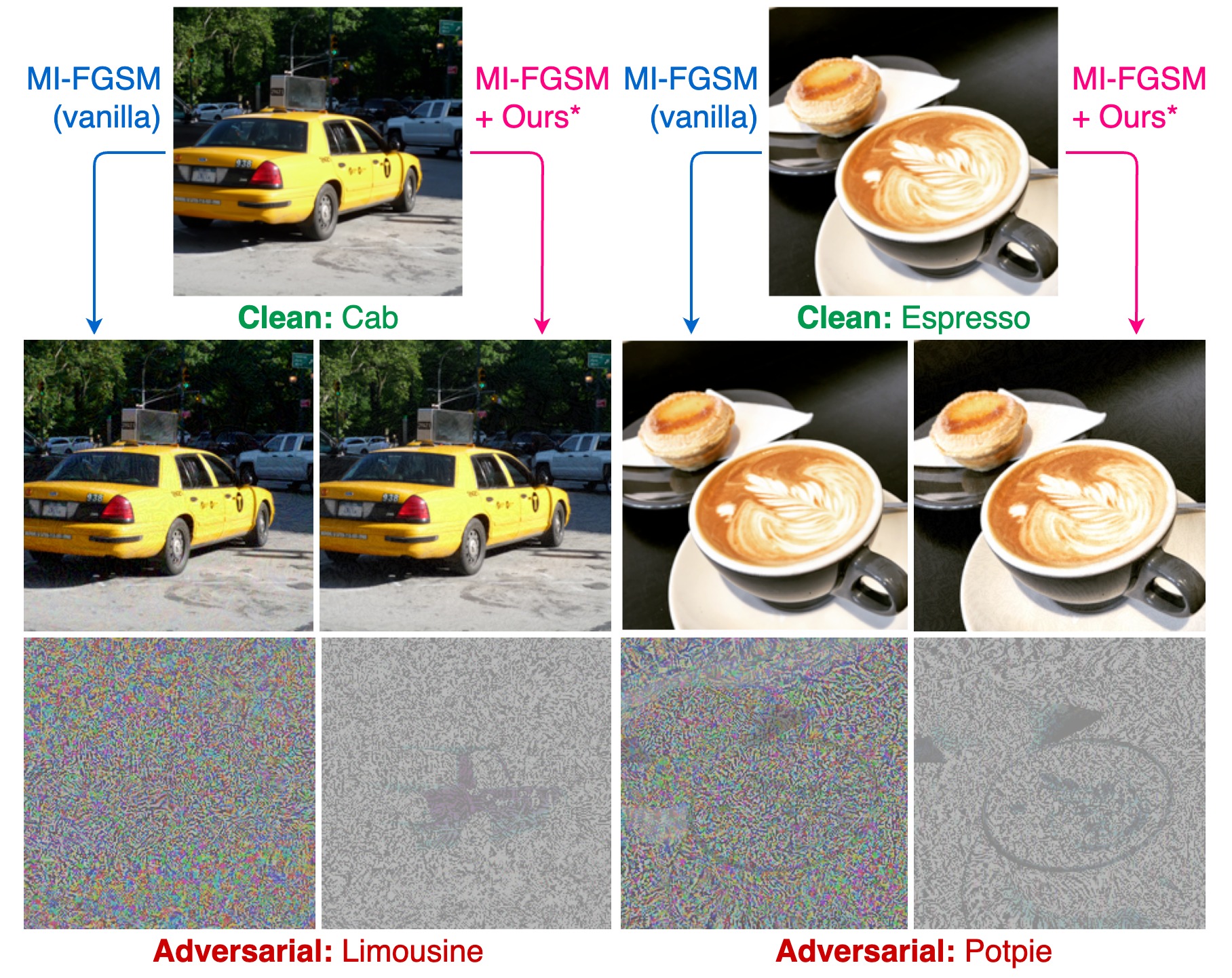}
    \caption{Adversarial examples crafted by vanilla MI-FGSM (blue arrow, left sample) and MI-FGSM boosted by our approach (pink arrow, right sample) on source model ResNet50 with $\varepsilon=8/255$ and $T=10$. Perturbation normalized for visualization (bottom row).}\label{fig:teaser}
\end{figure}

\citet{YaoGXKM19} and \citet{XuLZCZFEWL19} explored perturbation constraints within the spatial domain, boasting an improved efficiency for white-box attacks and effectiveness against model interpretability methods. While these attempts fall short for transfer-based black-box attacks, the idea of concentrating perturbation is valuable.

\citet{Ehrlich019} first showed intriguing properties of the frequency features of images that can be exploited. Prior work attempts to craft low-frequency perturbation with a fixed frequency constraint, assuming that the fixed low-frequency region reflect higher DNN sensitivity. However, DNNs do not respond uniformly nor unchangeably to these regions. Sensitivity to frequency coefficients varies across images, only with lower frequencies tending to be more impactful.
Our goal is to mitigate the effects of perturbation overfitting to the source model, to improve transferability. Motivated by this insight, we propose to strategically optimize adversarial perturbation by identifying sensitive frequency regions per input, per iteration, targeting influential frequency coefficients dynamically.
% Motivated by this, we intend to strategically manipulate adversarial perturbation, centralizing perturbation optimization dynamically towards dominant frequency regions, with an ultimate goal of mitigating the effect of perturbation's source model overfitting, thereby improving transferability.

In this paper, we propose to craft centralized adversarial perturbation, encompassing a shared frequency decomposition procedure, and two major strategies to regularize optimization. Frequency decomposition transforms data into frequency coefficient blocks with DCT (Discrete Cosine Transform). Perturbation centralization is performed via a quantization on these coefficients, omitting excessive perturbation, centralizing the optimization towards dominant regions. Our paramount contribution is a fine-grained quantization, controlled through a subsequent optimization of the quantization matrix, guaranteeing its direct alignment with regional sensitivity. Finally, we design our pipeline as a plug-and-play module, enabling seamless integration of our strategy into existing state-of-the-art gradient-based attacks.

Depicted in Figure~\ref{fig:teaser}, the effect of our proposed strategy to centralize perturbation is apparent. On the left side of the samples, perturbations generated by vanilla MI-FGSM exhibit obvious noise. In contrast, our centralized perturbation, while noticeably smaller in magnitude, yield equivalent adversarial potency. Our proposed optimization strategy eradicates model-specific perturbation, effectively avoiding source model overfitting. By purposely centralizing perturbation optimization within high-contribution frequency regions, improved adversarial effectiveness naturally arise. Our approach significantly enhances adversarial transferability by 11.7\% on average, and boosts attack efficacy on adversarially defended models by 10.5\% on average.

In summary, our contributions are as follows:

\begin{itemize}
    % \item We design a shared frequency decomposition process with DCT, transforming data into their frequency representations. Blocks of coefficients are quantized to leave out excessive perturbation, centralizing perturbation optimization, thereby avoiding source model overfitting.
    \item We design a shared frequency decomposition process with DCT. Excessive perturbation is eliminated through quantization on blocks of frequency coefficients. As such, adversarial perturbation is centralized, effectively avoiding source model overfitting.
    \item We propose a systematic approach for precise control over the quantization process. Quantization matrix is optimized dynamically in parallel with the attack, ensuring a direct alignment with model prediction at each step.
    \item Through comprehensive experiments, we substantiate the efficacy of our proposed centralized perturbation. Our approach yields significantly enhanced adversarial effectiveness. Under the same perturbation budget, centralized perturbation achieve stronger transferability, and is more successful in bypassing defenses.
\end{itemize}

\section{Related Work}\label{sec:related-work}

\paragraph{Adversarial Attacks.} Gradient-based attacks are extensively employed in both white-box and black-box transfer scenarios. The seminal work by \citet{GoodfellowSS14} introduced the single-step FGSM that exploits model gradients.~\citet{KurakinGB16} improves optimization with the multi-step iterative BIM. Numerous endeavors have been made to enhance the transferability of iterative gradient-based attacks, such as MI-FGSM~\cite{DongLPS0HL18}, TI-FGSM~\cite{DongPSZ19}, DI-FGSM~\cite{XieZZBWRY19}, SI-NI-FGSM~\cite{LinS00H20}, VMI-FGSM~\cite{Wang021}, among others. We use gradient-based iterative solutions as our attack basis.

\paragraph{Frequency Optimizations.}~\citet{Ehrlich019} first explored benefits of frequency-domain deep learning.~\citet{GuoFW19},~\citet{SharmaDB19}, and~\citet{DengFTU2020} exploit low-frequency image features to optimize perturbation within a predefined, constant low-frequency region. However, the assumption that DNN sensitivity maps to fixed frequency regions is disclaimed in~\citet{MaiyaFPAR21}, where different frequency regions yield varying sensitivity, with a higher sensitivity tendency towards lower coefficients. In contrast, our contribution lies in our effort to dynamically adjust frequency-domain perturbation optimization over back-propagation, augmenting attacks with centralized perturbation.

% In terms of frequency-enabled DNN vulnerabilities,~\citet{DuanCNYQH21} proposes to create adversarial examples by dropping information in the frequency domain,~\citet{LuoL0WXS22} leverages DWT (Discrete Wavelet Transform) to attack semantic similarity, and~\citet{LiZYL22} improves decision-based attacks via frequency mix-up. In contrast, our contribution lies in our effort to exploit the frequency sensitivity of DNNs, augmenting adversarial attacks with centralized perturbation.

\section{Methodology}
\newcommand{\bx}{\boldsymbol{x}}
\newcommand{\bxadv}{\bx^{adv}}
\newcommand{\bdelta}{\boldsymbol{\delta}}

\subsection{Overview}

Given a DNN classifier $\mathcal{F}(\cdot): \bx \in \mathbb{R}^m \mapsto y$, where $\bx$ and $y$ is the clean sample and ground truth label respectively. The adversary aims to create adversarial example $\bxadv = \bx + \bdelta$ with a minimal perturbation $\bdelta$ such that $\mathcal{F}(\bxadv) \neq y$, where $\bxadv$ is restricted by $\ell_p$-ball ($\ell_\infty$ in this work). Hence, the attack is an optimization which can be formulated as
\begin{equation}
    \label{eq:1}
    \arg\underset{\bxadv}{\max}\ \mathcal{J}(\bxadv, y),\ s.t.\ \|\bxadv -\bx\|_\infty\leqslant \varepsilon,
\end{equation}
where $\mathcal{J}(\cdot, \cdot)$ is the cross-entropy loss. Various gradient-based approaches have been proposed to solve this optimization as stated before, which we adopt as our foundation.

To optimize and craft centralized perturbation in the frequency domain, we devise a three-fold approach as follows:

\begin{enumerate}
    \item \textbf{Frequency coefficient decomposition:} Inspired by the JPEG codec, we design a shared differential data processing pipeline to decompose data into the frequency domain with DCT.
    \item \textbf{Centralized perturbation optimization:} Within the frequency domain, excessive perturbation is omitted via quantizing each Y/Cb/Cr channel, centralizing perturbation optimization.
    \item \textbf{Differential quantization optimization:} Quantization matrices are optimized through back-propagation, ensuring quantization aligns with dominant frequency regions.
\end{enumerate}

We detail these approaches in the following sections.

\subsection{Frequency Decomposition}

\begin{figure}[htbp]
    \centering
    \includegraphics[width=\linewidth]{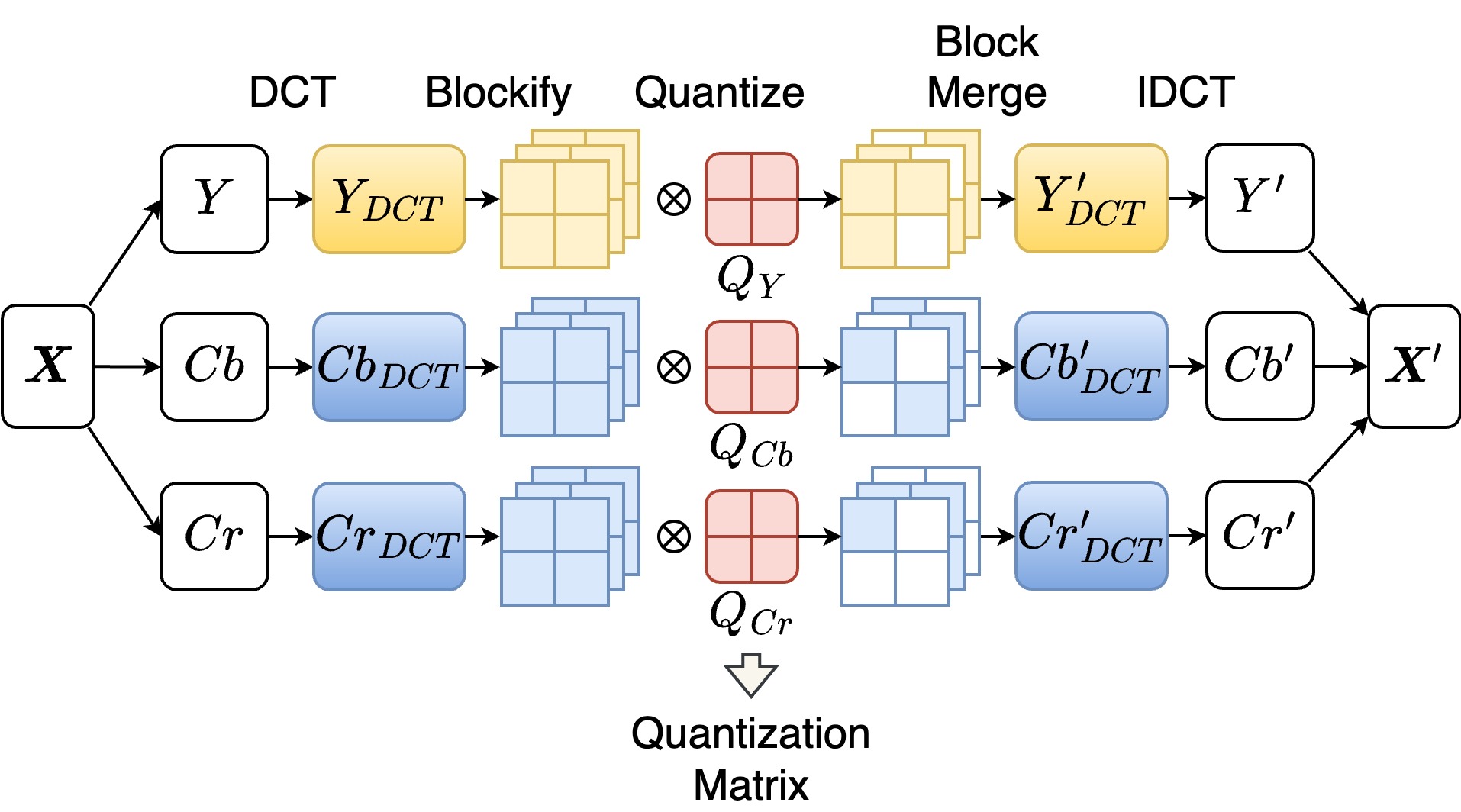}
    \caption{The frequency decomposition procedure.}\label{fig:dct-quantization}
\end{figure}

We begin by addressing the shared frequency decomposition procedure in our work. We focus on 8-bit RGB images of shape $(3, 224, 224)$. Illustrated in Figure~\ref{fig:dct-quantization}, let $\boldsymbol{X}$ be the data that we would like to \textit{frequency decompose}:

\begin{figure*}[t]
    \centering
    \includegraphics[width=0.88\textwidth]{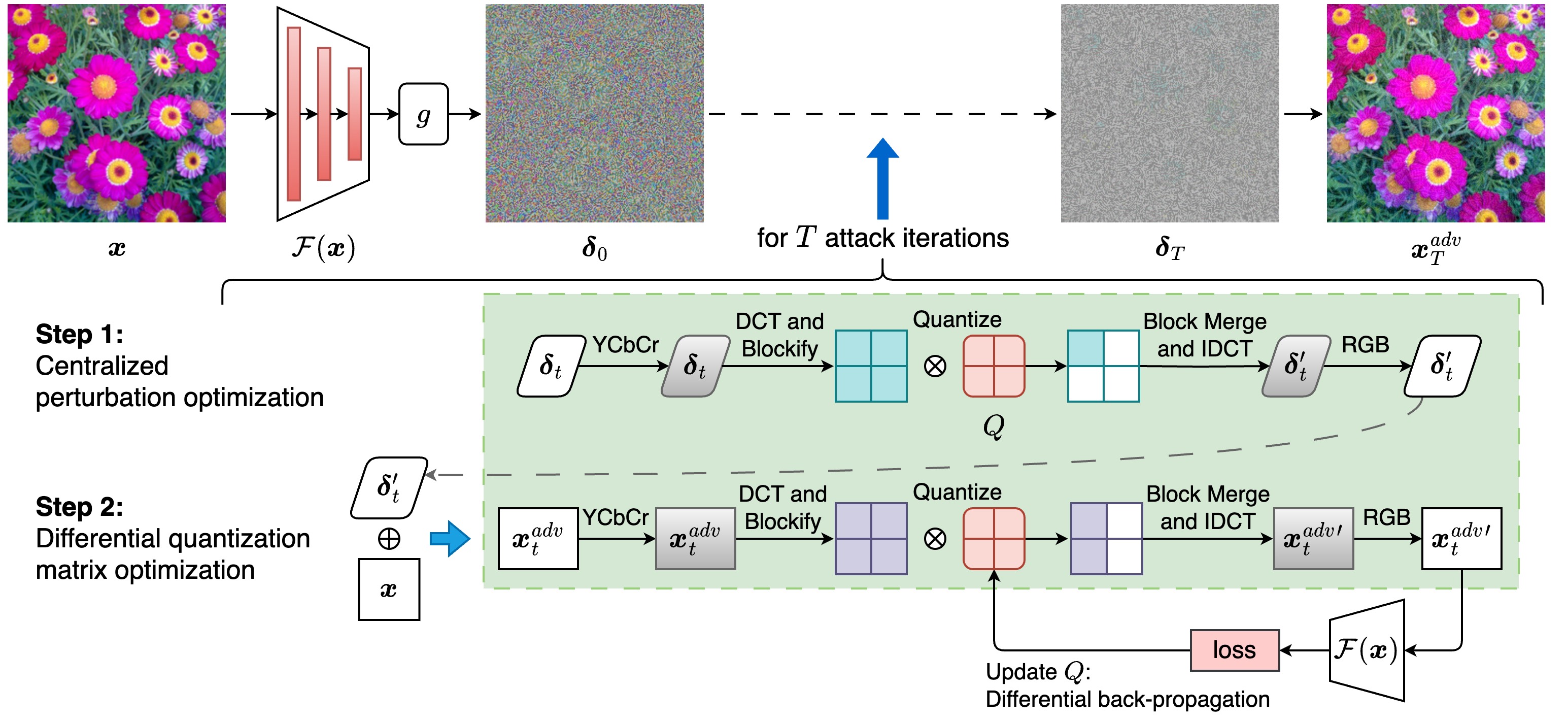}
    \caption{Pipeline of our strategy. Perturbation crafted with gradient-based attacks is centralized in the first step. Quantization matrix controlling the centralization is then optimized via back-propagation in the second step.}\label{fig:framework}
\end{figure*}

\begin{enumerate}
    \item We first convert $\boldsymbol{X}$ from RGB into the YCbCr color space, consisting of 3 color channels: the luma channel $Y$, and the chroma channels $Cb$ and $Cr$.
    \item After which, a channel-wise global DCT is applied to transform the data losslessly into the frequency domain.
    \item Next, a ``blockify'' process is used to reshape the data into blocks of $(8 \times 8)$. The operation is applied to the last two dimensions of the data (width $W$ and height $H$). As such, an image of shape $(B,C,W,H)$ would be ``blockified'' into shape $(B,C,W \cdot H/64,8,8)$ ($B$: batch, $C$: channel).
    \item Quantization is applied channel-wise via quantization matrices $Q$s, omitting excessive frequency coefficients.
    \item Finally, inverse operations, namely ``block merge'' and ``IDCT'' (inverse-DCT) is performed, reconstructing coefficients back into RGB image $\boldsymbol{X'}$.
\end{enumerate}

This procedure would decompose $\boldsymbol{X}$ into blocks of frequency coefficients in each Y/Cb/Cr channel. The sequential operation of ``blockify'' guarantees the lossless inversion of ``block merge''. Doing so, we ensure that every stage in our procedure is losslessly invertible, enabling reconstruction without information loss (with the exception of deliberate data quantization). The linearity of DCT and IDCT allow for the possibility of optimization within a reduced model space dimensionality, enabling the centralized perturbation and quantization to be optimized effectively.

This shared frequency decomposition procedure is shown in the green-bounded area in Figure~\ref{fig:framework}. For the next two sections, we address our optimization strategy after acquiring gradient-based attack crafted perturbation $\bdelta_t$ at iteration $t$.

\subsection{Centralized Perturbation Optimization}

Depicted in \textit{Step 1} of Figure~\ref{fig:framework}, first, the frequency decomposition is applied to $\bdelta_t$ at iteration $t$, where it is separated into luma and chroma channels, then decomposed into frequency coefficients via DCT and ``blockify''. Resulting components are denoted as $B_Y, B_{Cb}, B_{Cr}$, where
\begin{align}
    \begin{split}
        B_Y    & = \mathrm{blockify}(DCT(Y)),  \\
        B_{Cb} & = \mathrm{blockify}(DCT(Cb)), \\
        B_{Cr} & = \mathrm{blockify}(DCT(Cr)).
    \end{split}
\end{align}
Quantization matrices $Q_Y, Q_{Cb}, Q_{Cr}$ are applied over each Y/Cb/Cr channel as
\begin{align}
    \begin{split}
        B_Y'    & = B_Y \odot Q_Y,       \\
        B_{Cb}' & = B_{Cb} \odot Q_{Cb}, \\
        B_{Cr}' & = B_{Cr} \odot Q_{Cr}. \\
    \end{split}
\end{align}
Lastly, inverse operations are taken as
\begin{align}
    \begin{split}
        Y'  & = IDCT(\mathrm{block\mbox{-}merge}(B'_Y)),    \\
        Cb' & = IDCT(\mathrm{block\mbox{-}merge}(B'_{Cb})), \\
        Cr' & = IDCT(\mathrm{block\mbox{-}merge}(B'_{Cr})),
    \end{split}
\end{align}
and with an RGB conversion, reconstruction is complete.

Differing from the quantization matrix employed in the JPEG codec, ours is defined as $Q = (q_{ij}) \in \{0,1\} ^{m \times m}$ where $m=8$, and initialized as $Q_0 = \mathds{1}$. Through the block-wise multiplication of $B \odot Q$, less impactful frequency coefficients are systematically discarded at every iteration. This process confines the perturbation optimization exclusively within frequency regions of greater impact over DNN predictions, as modeled by the pattern in $Q$.

At iteration $t$, perturbation $\bdelta_t$ will be centralized via optimization, yielding $\bdelta_t'$. We denote the entire process as
\begin{equation}
    \label{eq:4}
    \bdelta_t'=\mathcal{K}(\bdelta_t; Q_t),
\end{equation}
where $\mathcal{K}(\cdot; Q)$ is the frequency decomposition and quantization function with quantization matrix $Q$. $Q_t$ is fixed at this stage per each iteration $t$.

\subsection{Differential Quantization Matrix Optimization}

Intuitively, quantization matrix $Q$ holds pivotal significance in the process of centralizing frequency-sensitive perturbation. In this section, we delve into the modeling and optimization of $Q$ to ensure its direct alignment with frequency-wise coefficient influence on DNN predictions.

Illustrated in \textit{Step 2} of Figure~\ref{fig:framework}, during this stage, the optimized $\bdelta_t'$ is added onto $\bx$ to get intermediate adversarial example $\bxadv_t$ at iteration $t$. $\bxadv_t$ goes through the identical shared frequency decomposition process, and is quantized in the same manner as \textit{Step 1}, giving us $\bxadv_t{'}$. The quantized $\bxadv_t{'}$ is then fed back into source model $\mathcal{F}(\cdot)$, such that $Q$ is updated subsequently through back-propagation.

We formulate the update of $Q$ as an optimization problem by leveraging the gradient changes of the source model before and after the quantization of $\bxadv_t$. Our optimization goal is: $Q$ should quantize $\bxadv_t$ in a way that, at each iteration, the source model should be less convinced that $\bxadv_t$ is $y$. As such, the loss is formulated as
\begin{equation}
    \label{eq:5}
    \arg\underset{Q_t}{\max}\ \mathcal{J}(\mathcal{K}(\bxadv_t; Q_t), y),
\end{equation}
where $Q_t$ is the variable for optimization. $Q$ consists of $Q_Y, Q_{Cb}, Q_{Cr}$ for each Y/Cb/Cr channel respectively, and is collectively optimized by the Adam optimizer to solve this optimization. With this approach, we assure that (1) $Q$ accurately reflects the impact of frequency coefficients on model predictions \textit{per each iteration}, allowing for fine-grained control over the centralized perturbation quantization process, and (2) gradients from the source model accumulate across successive iterations, further boosting transferability.

Through the process of back-propagation, $Q$ undergoes iterative updates via optimization. We denote the optimized result as matrix $P=(p_{ij}) \in \mathbb{R}^{m\times m}$. Following the update, a rounding function $\mathcal{R}(\cdot)$ is implemented before $Q$ is applied to the quantization process. Specifically, for a quantization ratio of $r$ where $0 \leqslant r \leqslant 1$ for each Y/Cb/Cr channel,
\begin{equation}
    \label{eq:6}
    Q=\mathcal{R}(P;r)=
    \begin{cases}
        1,\quad \mathrm{where}\ p_{ij} \geqslant \rho, \\
        0,\quad \mathrm{otherwise},                    \\
    \end{cases}
\end{equation}
where $\rho$ is the $r$-th percentile of $\{ p_{ij} \}$, i.e., the threshold for each Y/Cb/Cr channel. While Equation~\ref{eq:6} is represented as a staircase (i.e., non-differentiable), $\mathcal{R}(\cdot)$ is implemented in a differentiable manner, which we detail in the Appendix.

\newcommand{\bg}{\boldsymbol{g}}

\begin{algorithm}[tb]
    \caption{Centralized Adversarial Perturbation}%
    \label{alg:the-algorithm}
    \textbf{Input}: Original image $\bx$, ground truth label $y$, source model $\mathcal{F}$ with loss $\mathcal{J}$.\\
    \textbf{Parameters}: Iteration $T$, size of perturbation $\varepsilon$, learning rate $\beta$, quantization ratios $r_Y, r_{Cb}, r_{Cr}$ and corresponding quantization matrices $Q_Y, Q_{Cb}, Q_{Cr}$ (denoted as $Q$s for brevity) each Y/Cb/Cr channel. \\
    \textbf{Output}: $\bxadv_T$.
    \begin{algorithmic}[1] %[1] enables line numbers
        \STATE{Let step size $\alpha \leftarrow \varepsilon / T$, $Q_0=\mathds{1}$.}
        \FOR{$t=0$ \textbf{to} $T-1$}
        \STATE{Acquire gradient from $\mathcal{F}$ as $\nabla_{\bx} \mathcal{J} (\bxadv_t, y)$.}
        \STATE{Perturbation $\bdelta_t = \alpha \cdot \mathrm{sign}(\nabla_{\bx} \mathcal{J} (\bxadv_t, y)) $.}
        \STATE{Optimize $\bdelta_t$ by Equation~\ref{eq:4} and~\ref{eq:6} to acquire $\bdelta_t'$, and clip with respect to $\varepsilon$.}
        \STATE{Acquire intermediate $\bxadv_t=\bx+\bdelta_t'$.}
        \STATE{Update $Q_t$s by passing the quantized $\bxadv_t$ (through the same Equation~\ref{eq:4} and~\ref{eq:6}) to $\mathcal{F}$ and solving the optimization via Equation~\ref{eq:5}.}
        \ENDFOR%
        \STATE{\textbf{return} $\bxadv_T = \bx +\bdelta_T'$.}
    \end{algorithmic}
\end{algorithm}

\subsection{The Algorithm}

Finally, Algorithm~\ref{alg:the-algorithm} showcase a basic attack integrated with our strategy, crafting centralized adversarial perturbation (based on BIM). After the conventional step of creating intermediate perturbation from the source model's gradients, our approach of centralizing the perturbation in a frequency-sensitive manner, and updating the quantization matrix with respect to model predictions, is a streamline integration.

Our strategy guarantees that the centralized perturbation through frequency quantization is concentrated into frequency regions that hold greater significant to the model's judgements. Updating the quantization matrix per iteration allows for a more fine-grained control, making sure that each optimization aligns with the iteration-local gradients of the model. Lastly, as a bonus, at each iteration, gradients gets carried over and accumulated to the next through the directional optimization of the quantization matrix, improving adversarial transferability.

\section{Experiments}

\subsection{Setup}

\begin{figure*}[t]
    \centering
    \includegraphics[width=\linewidth]{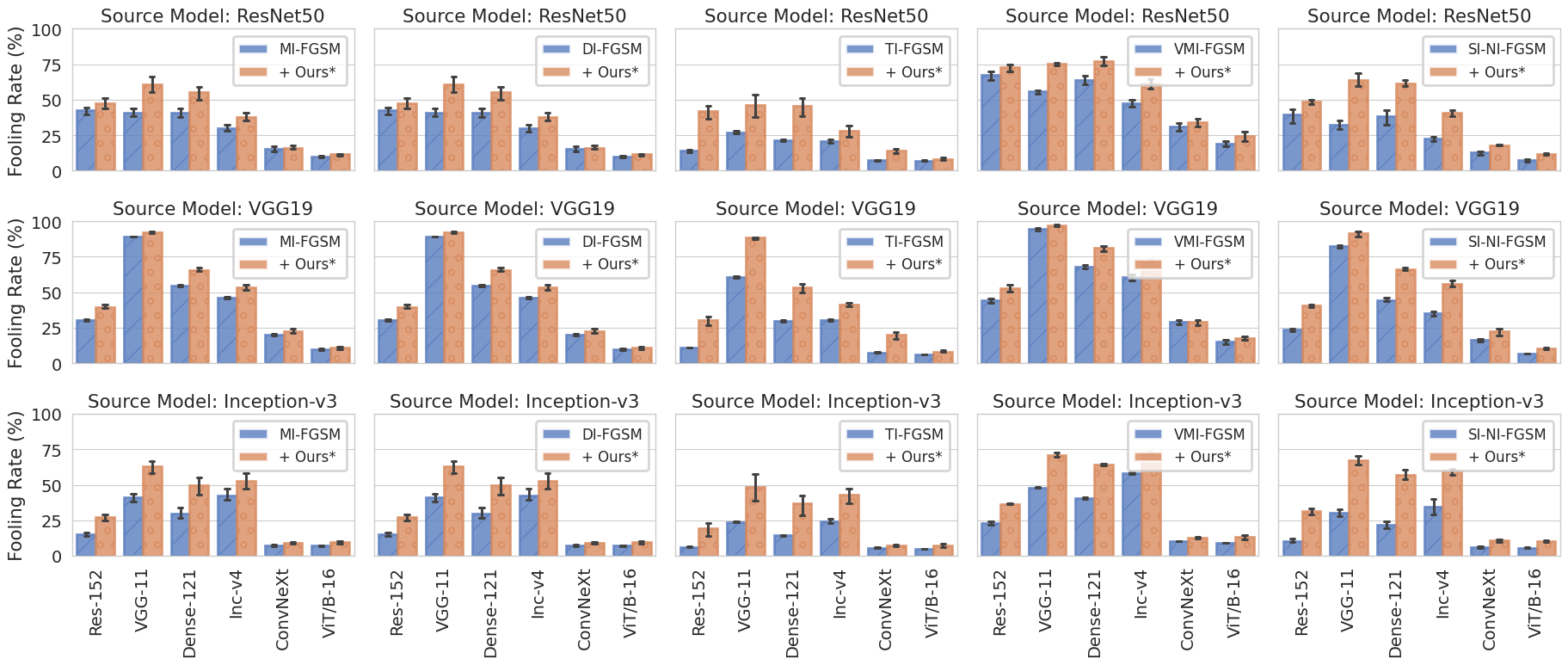}
    \caption{Evaluation of adversarial transferability. Transfer fooling rates are aggregated across attacks over $T$.}\label{fig:transfer-eval}
\end{figure*}
\paragraph{Dataset.} The dataset from the NeurIPS 2017 Adversarial Learning Challenge~\cite{AlexyAADC2018} is used, consisting of 1000 images from ImageNet with shape $(3, 224, 224)$.

\paragraph{Networks.} We choose 3 source models as local surrogate models: ResNet-50 (Res-50)~\cite{HeZRS16}, VGG-19~\cite{SimonyanZ14a}, Inception-v3 (Inc-v3)~\cite{SzegedyVISW16}. All pretrained models are sourced from~\citet{rw2019timm}.

\paragraph{Attacks.} Our approach is integrated and evaluated against 5 state-of-the-art gradient-based attacks, namely MI-FGSM, DI-FGSM, TI-FGSM, VMI-FGSM, and SI-NI-FGSM, for a better presentation of transferability. The unmodified forms of these attacks serve as the baselines for our experiments.

\paragraph{Implementation Details.} We intentionally drop excessive perturbation, centralizing optimization towards dominant frequency regions. As such, to enable a fair comparison between our attack and baseline methods: for all three Y/Cb/Cr channels, each set at a quantization ratio of $r_Y, r_{Cb}, r_{Cr}$, the cumulative quantization rate is $(r_Y+r_{Cr}+r_{Cr}) / 3$; With a baseline attack under an $\ell_\infty$ constraint of $\varepsilon_0$, we equate our attack to the same perturbation budget by imposing a constraint of $\varepsilon = \varepsilon_0 / ((r_Y+r_{Cr}+r_{Cr}) / 3)$.

\paragraph{Hyper-parameters.} Attacks are $\ell_\infty$-bounded. $\varepsilon=8/255$. They run for iteration $T=10, 20, 50$. Learning rate of the Adam optimizer $\beta=0.1$. Quantization ratios are set to $r_Y=0.9, r_{Cb}=0.05, r_{Cr}=0.05$, with a total quantization rate of $1/3$. All other parameters are kept exact to their original implementations.

\subsection{Transferability}

We initiate our evaluation with our primary focus: adversarial transferability. 6 normally trained models: ResNet-152 (Res-152), VGG-11, DenseNet-121 (Dense-121)~\cite{HuangLMW17}, Inception-v4 (Inc-v4)~\cite{SzegedyIVA17}, ConvNeXt~\cite{0003MWFDX22}, and ViT-B/16~\cite{DosovitskiyB0WZ21} are used as black-box target models.

Figure~\ref{fig:transfer-eval} presents the transfer fooling rates of the crafted adversarial examples used to attack the black-box models. Fooling rates are averaged over parameter: attack iteration $T$. We report a consistent improvement in transfer fooling rates when integrating our proposed frequency perturbation centralization strategy with vanilla gradient-based attacks. Displayed in orange, our approach achieves an average boost in adversarial transferability of 11.7\% compared to the baseline attacks represented in blue.

We argue that the observed increase in transferability is the direct manifestation of the benefits that centralized perturbation brings in our method. Dominating frequency features of an image is learnt and shared by neural networks alike, while excessive perturbation that are crafted to fit model-specific features is what hinders the transferability of adversarial examples. By centralizing adversarial perturbation towards the shared dominant frequency regions, we effectively disrupt the model's judgment in a more generalized manner, boosting adversarial transferability.

\begin{figure*}[!t]
    \centering
    \includegraphics[width=\linewidth]{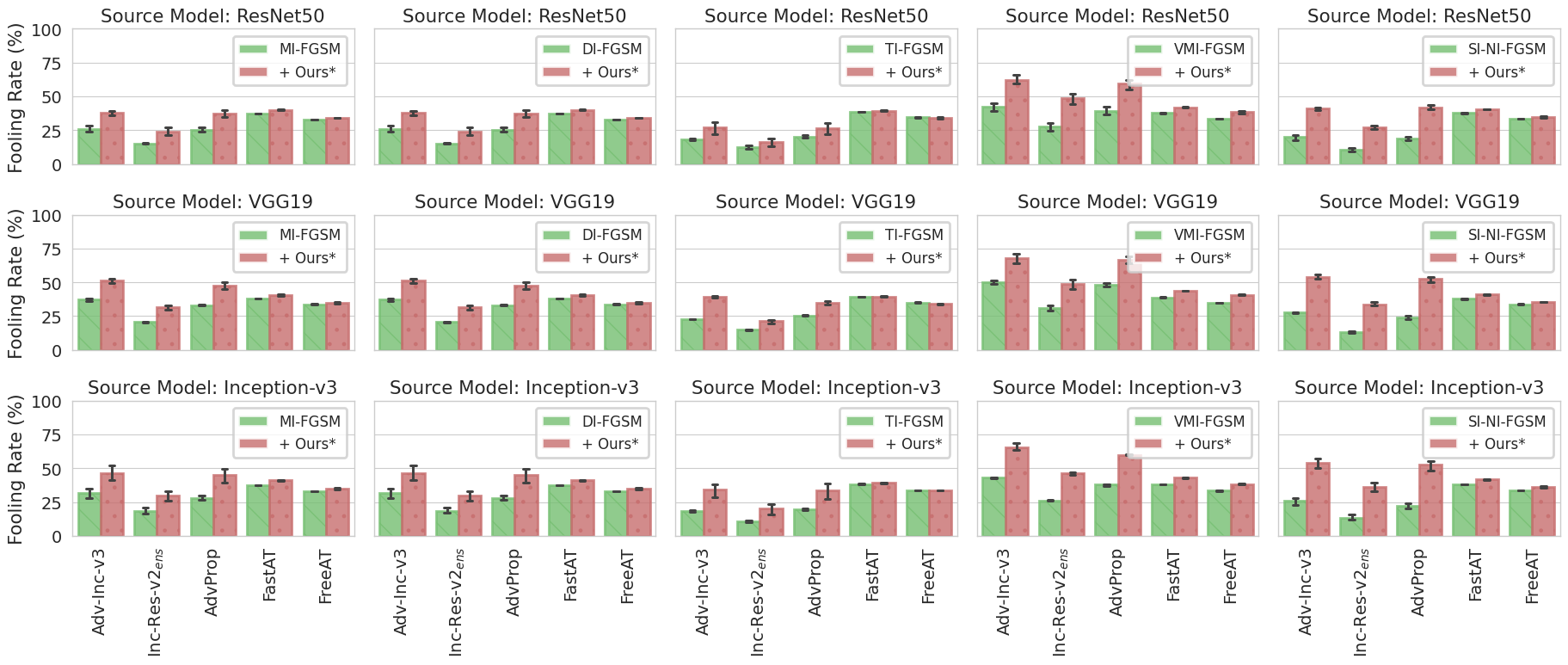}
    \caption{Evaluation of adversarial effectiveness against adversarial training-based defenses. Fooling rates aggregated over $T$.}\label{fig:defense-eval}
\end{figure*}

\subsection{Defense Evasion}

Currently two types of defenses provide an extra layer of robustness guarantee to neural networks: (1) filter-based defenses, where perturbation is filtered through image transformations, and (2) adversarial training, where models are strengthened by training on adversarial examples.

\begin{table}[!t]
    \centering
    \resizebox{\linewidth}{!}{
        \begin{tabular}{m{1.1cm}m{0.9cm}m{0.9cm}m{0.9cm}m{0.9cm}m{0.9cm}m{0.9cm}m{0.9cm}m{0.9cm}}
            \toprule
            \textbf{Attack}   & \textbf{Res-152}       & \textbf{VGG-11}         & \textbf{Dense-121}     & \textbf{Inc-v4}        & \textbf{Conv-NeXt}     & \textbf{ViT-B/16}      \\ \midrule
            \multicolumn{7}{c}{\textbf{JPEG compression --- quality level 75}}                                                                                                       \\ \midrule
            \textbf{MI-FGSM}  & 23.1\% (±2.0)          & 59.2\% (±13.8)          & 41.3\% (±5.7)          & 35.0\% (±6.7)          & 15.3\% (±1.3)          & 12.3\% (±0.5)          \\
            \rowcolor{gray!20}
            \textbf{+Ours*}   & \textbf{34.8\%} (±1.3) & \textbf{74.8\%} (±11.0) & \textbf{55.8\%} (±5.3) & \textbf{47.2\%} (±6.2) & \textbf{22.3\%} (±1.9) & \textbf{16.3\%} (±1.0) \\
            \textbf{VMI-FGSM} & 33.1\% (±6.0)          & 66.3\% (±15.3)          & 51.7\% (±6.8)          & 45.7\% (±7.3)          & 22.0\% (±2.8)          & 17.0\% (±1.5)          \\
            \rowcolor{gray!20}
            \textbf{+Ours*}   & \textbf{48.1\%} (±5.7) & \textbf{81.0\%} (±11.8) & \textbf{70.4\%} (±5.0) & \textbf{62.8\%} (±5.9) & \textbf{31.8\%} (±3.5) & \textbf{23.8\%} (±3.7) \\
            \midrule
            \multicolumn{7}{c}{\textbf{Bit-depth reduction --- to 3 bits}}                                                                                                           \\ \midrule
            \textbf{MI-FGSM}  & 17.9\% (±2.8)          & 57.7\% (±12.4)          & 37.8\% (±6.2)          & 33.3\% (±5.5)          & 14.3\% (±0.9)          & 10.4\% (±0.3)          \\
            \rowcolor{gray!20}
            \textbf{+Ours*}   & \textbf{30.3\%} (±0.8) & \textbf{71.6\%} (±11.1) & \textbf{53.5\%} (±6.0) & \textbf{43.3\%} (±7.3) & \textbf{18.4\%} (±1.4) & \textbf{13.9\%} (±1.4) \\
            \textbf{VMI-FGSM} & 28.8\% (±8.1)          & 63.9\% (±14.4)          & 47.3\% (±6.7)          & 42.3\% (±7.4)          & 20.5\% (±2.8)          & 13.5\% (±1.5)          \\
            \rowcolor{gray!20}
            \textbf{+Ours*}   & \textbf{44.7\%} (±8.7) & \textbf{78.0\%} (±12.5) & \textbf{69.0\%} (±4.9) & \textbf{57.8\%} (±6.8) & \textbf{27.0\%} (±3.0) & \textbf{20.3\%} (±2.5) \\
            \bottomrule
        \end{tabular}}
    \caption{Evaluation of adversarial effectiveness against filter-based adversarial defenses. Fooling rates aggregated with $\mathrm{mean}$s and $\mathrm{std}$s over attack iteration $T$.}\label{tab:transform-defense}
\end{table}

\begin{figure}[!t]
    \centering
    \includegraphics[width=\linewidth]{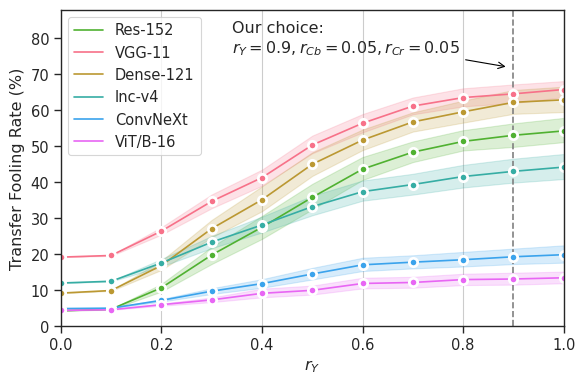}
    \caption{The effect of the quantization ratio of luma channel $r_Y$ vs. chroma channels $r_{Cb}, r_{Cr}$ over transferability.}\label{fig:effect-of-r}
\end{figure}

\begin{figure*}[t]
    \centering
    \includegraphics[width=\linewidth]{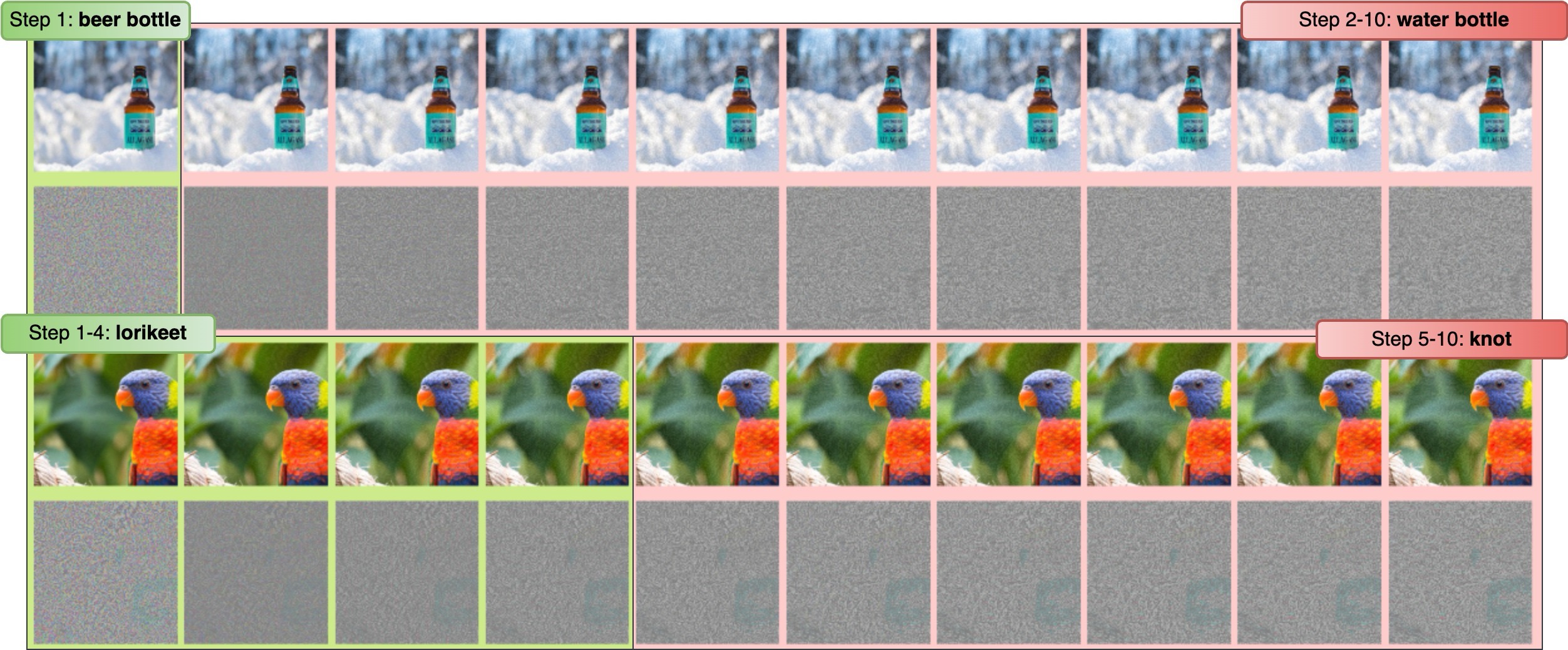}
    \caption{A visualization of perturbation over $T=10$ optimization iterations (from left to right) of MI-FGSM attacking ResNet50. Correct predictions are framed in green, where incorrect ones are framed in red.}\label{fig:perturb-over-iters}
\end{figure*}

\paragraph{Filter-based Defense.}
\citet{GuoRCM18} first explored using JPEG compression and bit-depth reduction as defenses against adversarial attacks. Since our approach also utilizes similar techniques (such as drawing inspiration from the JPEG codec for perturbation quantization), we are interested in investigating how our method withstands these defenses.

Shown in Table~\ref{tab:transform-defense}, we report the transfer fooling rates of attacks: MI-FGSM, VMI-FGSM and variants integrated with our approach. Highlighted in bold, we observe that attacks integrated with our strategy all outperform their baselines, with an increase of at least 3.5\%. We posit that these defenses fail as they intend to preserve dominant image features to maintain visual quality, assuming that perturbation lies within unimportant regions to ensure imperceptibility. As such, they are unable to remove our perturbation designed to centralize around dominating regions, enabling our strategy to bypass these defenses.

\paragraph{Adversarial Training.}
5 adversarially trained black-box models: Adv-Inc-v3, Inc-Res-v2$_{ens}$~\cite{TramerKPGBM18}, Adv-ResNet-50 (FastAT), Adv-ResNet-50 (FreeAT)~\cite{ShafahiNG0DSDTG19}, and EfficientNet-B0 (AdvProp)~\cite{XieTGWYL20} are used for evaluating our approach against adversarial training as a defense mechanism.

We report our results in Figure~\ref{fig:defense-eval}, where we observe a steady boost of fooling rates by a maximum of 20\% in some scenarios by our strategy. In all, results indicate that our proposed approach can successfully evade adversarial defenses that transfer-based attacks frequently fail against.

\subsection{The Impact of Quantization Ratio Allocation}

The quantization process is crucial for centralized perturbation optimization, controlled by quantization ratios $r$s. We explain our rationale for the choice of ratios through evaluating the impact of luma channel $r_Y$ vs.\ chroma channels $r_{Cb},r_{Cr}$ on adversarial effectiveness.

We maintain a cumulative quantization rate of $1/3$, iterating the luma channel $r_Y$ from $0$ to $1$ with step size $0.1$, and allocating the rest equally to the chroma channels $r_{Cb}$ and $r_{Cr}$. Figure~\ref{fig:effect-of-r} shows the transfer fooling rates of MI-FGSM integrated with our approach. As $r_Y$ increases, the fooling rates show a consistent increase as well. We reason that this occurs because the luma channel contains more structural information, which DNNs commonly learn as more useful features compared to the chroma channels. However, changes in the luma channel are also more noticeable by the human visual system. As such, we choose $r_Y=0.9$ for a balance of adversarial effectiveness and perturbation imperceptibility. Further investigation is addressed in the Appendix.

\subsection{Visualization of Perturbation Optimization}

In Figure~\ref{fig:perturb-over-iters}, we visualize the perturbation optimized by MI-FGSM + our approach over 10 iterations, giving further insight to the centralized optimization process of our strategy.

On iteration 1 (step $t=1$), quantization is initialized as $Q_0=\mathds{1}$. On $t=2$, we observe an immediate drop in perturbation, indicating quantization is applied. Going further ($t=3$ to $10$), we find perturbation is gradually added at each iteration, centralized on more significant frequency regions. By ensuring a consistent optimization alignment with model judgement, our centralized perturbation succeeds in exploiting dominating frequency features of an image.

\subsection{Ablation Study}

We conduct an ablation study on our proposed centralized perturbation optimization process --- the core of our strategy. We consider 4 other strategies: (1) \textit{RandA}: randomly choose $Q$s fixed at the start, (2) \textit{RandB}: randomly choose $Q$s at each iteration, (3) \textit{Low}: preserving only low-frequency coefficients (which is what previous literature proposes to do), and (4) \textit{High}: preserving only high-frequency coefficients. Details addressed in the Appendix.

In Table~\ref{tab:ablation}, we report the fooling rates of vanilla MI-FGSM and SI-NI-FGSM, and ones integrated with \textit{RandA}, \textit{RandB}, \textit{Low}, and our strategy.\ (\textit{High} is ignored as fooling rates mostly dropped to zero when it is integrated.) Highlighted in bold, we prove our strategy's ability as our approach consistently boosts transferability. In contrast, \textit{RandA} and \textit{RandB} both fail to achieve comparable adversarial effectiveness even with the baselines.
While \textit{Low} comes close to matching and occasionally surpasses us, our approach still maintains the upper hand across the majority of cases. We argue that \textit{Low} and our approach both acknowledges that the low-frequency regions hold more dominating features. However, instead of brute-forcely constraining all perturbation towards consecutive low-frequency coefficients, our strategy's precise control over the centralization process yields an improved adversarial effectiveness.

\begin{table}[!t]
    \centering
    \resizebox{\linewidth}{!}{
        \begin{tabular}{m{0.9cm}m{0.8cm}m{0.8cm}m{0.8cm}m{0.8cm}m{0.8cm}m{0.8cm}m{0.8cm}}
            \toprule
            \textbf{Variant} & \textbf{White-box} & \textbf{Res-152} & \textbf{VGG-11} & \textbf{Dense-121} & \textbf{Inc-v4} & \textbf{Conv-NeXt} & \textbf{ViT-B/16} \\ \midrule
            \multicolumn{8}{c}{\textbf{MI-FGSM}}                                                                                                                       \\ \midrule
            \textbf{Vanilla} & 100\%              & 44.6\%           & 43.0\%          & 45.3\%             & 32.0\%          & 16.6\%             & 10.3\%            \\
            \textbf{RandA}   & 86.2\%             & 37.0\%           & 52.1\%          & 45.2\%             & 31.4\%          & 13.0\%             & 9.4\%             \\
            \textbf{RandB}   & 92.3\%             & 37.6\%           & 52.2\%          & 45.0\%             & 31.8\%          & 13.7\%             & 9.6\%             \\
            \textbf{Low}     & 100\%              & 50.3\%           & 63.1\%          & 56.2\%             & 39.5\%          & 18.9\%             & 11.9\%            \\
            \rowcolor{gray!20}
            \textbf{Ours*}   & 100\%              & \textbf{50.7\%}  & \textbf{65.7\%} & \textbf{58.7\%}    & \textbf{41.0\%} & 17.8\%             & \textbf{12.0\%}   \\ \midrule
            \multicolumn{8}{c}{\textbf{SI-NI-FGSM}}                                                                                                                    \\ \midrule
            \textbf{Vanilla} & 99.9\%             & 43.2\%           & 35.4\%          & 42.9\%             & 24.1\%          & 13.4\%             & 8.0\%             \\
            \textbf{RandA}   & 88.4\%             & 36.7\%           & 61.5\%          & 49.3\%             & 34.8\%          & 14.7\%             & 10.1\%            \\
            \textbf{RandB}   & 76.6\%             & 28.2\%           & 51.4\%          & 42.0\%             & 29.1\%          & 11.6\%             & 8.7\%             \\
            \textbf{Low}     & 97.4\%             & 45.5\%           & 66.5\%          & 57.9\%             & 41.2\%          & 18.6\%             & 11.9\%            \\
            \rowcolor{gray!20}
            \textbf{Ours*}   & \textbf{98.7\%}    & \textbf{49.9\%}  & \textbf{68.9\%} & \textbf{64.1\%}    & \textbf{42.6\%} & 17.6\%             & 11.2\%            \\
            \bottomrule
        \end{tabular}}
    \caption{Ablation study of the centralized perturbation optimization strategy. Source model chosen as ResNet50.}\label{tab:ablation}
\end{table}

\section{Conclusion}

We propose a novel approach to centralize adversarial perturbation to dominant frequency regions. Our technique involves dynamic perturbation optimization through frequency domain quantization, resulting in centralized perturbation that offers improved generalization. To ensure precise quantization alignment with the model's judgement, we incorporate a parallel optimization via back-propagation. Experiments prove that our strategy achieves remarkable transferability and succeeds in evading various defenses. Implications of our work prompt further exploration of defending centralized perturbation under a frequency context.

\section*{Acknowledgements}

This work was supported by the National Natural Science Foundation of China (Grants No. 62072037 and U1936218).

\bibliography{aaai24}

\begin{thebibliography}{31}
\providecommand{\natexlab}[1]{#1}

\bibitem[{Deng and Karam(2020)}]{DengFTU2020}
Deng, Y.; and Karam, L.~J. 2020.
\newblock Frequency-Tuned Universal Adversarial Attacks.
\newblock \emph{CoRR}, abs/2003.05549.

\bibitem[{Dong et~al.(2018)Dong, Liao, Pang, Su, Zhu, Hu, and Li}]{DongLPS0HL18}
Dong, Y.; Liao, F.; Pang, T.; Su, H.; Zhu, J.; Hu, X.; and Li, J. 2018.
\newblock Boosting Adversarial Attacks With Momentum.
\newblock In \emph{{CVPR}}, 9185--9193. Computer Vision Foundation / {IEEE} Computer Society.

\bibitem[{Dong et~al.(2019)Dong, Pang, Su, and Zhu}]{DongPSZ19}
Dong, Y.; Pang, T.; Su, H.; and Zhu, J. 2019.
\newblock Evading Defenses to Transferable Adversarial Examples by Translation-Invariant Attacks.
\newblock In \emph{{CVPR}}, 4312--4321. Computer Vision Foundation / {IEEE}.

\bibitem[{Dosovitskiy et~al.(2021)Dosovitskiy, Beyer, Kolesnikov, Weissenborn, Zhai, Unterthiner, Dehghani, Minderer, Heigold, Gelly, Uszkoreit, and Houlsby}]{DosovitskiyB0WZ21}
Dosovitskiy, A.; Beyer, L.; Kolesnikov, A.; Weissenborn, D.; Zhai, X.; Unterthiner, T.; Dehghani, M.; Minderer, M.; Heigold, G.; Gelly, S.; Uszkoreit, J.; and Houlsby, N. 2021.
\newblock An Image is Worth 16x16 Words: Transformers for Image Recognition at Scale.
\newblock In \emph{{ICLR}}. OpenReview.net.

\bibitem[{Ehrlich and Davis(2019)}]{Ehrlich019}
Ehrlich, M.; and Davis, L. 2019.
\newblock Deep Residual Learning in the {JPEG} Transform Domain.
\newblock In \emph{{ICCV}}, 3483--3492. {IEEE}.

\bibitem[{Goodfellow, Shlens, and Szegedy(2015)}]{GoodfellowSS14}
Goodfellow, I.~J.; Shlens, J.; and Szegedy, C. 2015.
\newblock Explaining and Harnessing Adversarial Examples.
\newblock In \emph{{ICLR} (Poster)}.

\bibitem[{Guo, Frank, and Weinberger(2019)}]{GuoFW19}
Guo, C.; Frank, J.~S.; and Weinberger, K.~Q. 2019.
\newblock Low Frequency Adversarial Perturbation.
\newblock In \emph{{UAI}}, volume 115 of \emph{Proceedings of Machine Learning Research}, 1127--1137. {AUAI} Press.

\bibitem[{Guo et~al.(2018)Guo, Rana, Ciss{\'{e}}, and van~der Maaten}]{GuoRCM18}
Guo, C.; Rana, M.; Ciss{\'{e}}, M.; and van~der Maaten, L. 2018.
\newblock Countering Adversarial Images using Input Transformations.
\newblock In \emph{{ICLR} (Poster)}. OpenReview.net.

\bibitem[{Guo et~al.(2019)Guo, Zhao, Li, Kuang, Zhang, Han, and Tan}]{GuoZLKZHT19}
Guo, F.; Zhao, Q.; Li, X.; Kuang, X.; Zhang, J.; Han, Y.; and Tan, Y. 2019.
\newblock Detecting adversarial examples via prediction difference for deep neural networks.
\newblock \emph{Inf. Sci.}, 501: 182--192.

\bibitem[{He et~al.(2016)He, Zhang, Ren, and Sun}]{HeZRS16}
He, K.; Zhang, X.; Ren, S.; and Sun, J. 2016.
\newblock Deep Residual Learning for Image Recognition.
\newblock In \emph{{CVPR}}, 770--778. {IEEE} Computer Society.

\bibitem[{Huang et~al.(2017)Huang, Liu, van~der Maaten, and Weinberger}]{HuangLMW17}
Huang, G.; Liu, Z.; van~der Maaten, L.; and Weinberger, K.~Q. 2017.
\newblock Densely Connected Convolutional Networks.
\newblock In \emph{{CVPR}}, 2261--2269. {IEEE} Computer Society.

\bibitem[{Kurakin, Goodfellow, and Bengio(2016)}]{KurakinGB16}
Kurakin, A.; Goodfellow, I.~J.; and Bengio, S. 2016.
\newblock Adversarial examples in the physical world.
\newblock \emph{CoRR}, abs/1607.02533.

\bibitem[{Kurakin et~al.(2018)Kurakin, Goodfellow, Bengio, Dong, Liao, Liang, Pang, Zhu, Hu, Xie, Wang, Zhang, Ren, Yuille, Huang, Zhao, Zhao, Han, Long, Berdibekov, Akiba, Tokui, and Abe}]{AlexyAADC2018}
Kurakin, A.; Goodfellow, I.~J.; Bengio, S.; Dong, Y.; Liao, F.; Liang, M.; Pang, T.; Zhu, J.; Hu, X.; Xie, C.; Wang, J.; Zhang, Z.; Ren, Z.; Yuille, A.~L.; Huang, S.; Zhao, Y.; Zhao, Y.; Han, Z.; Long, J.; Berdibekov, Y.; Akiba, T.; Tokui, S.; and Abe, M. 2018.
\newblock Adversarial Attacks and Defences Competition.
\newblock \emph{CoRR}, abs/1804.00097.

\bibitem[{Lin et~al.(2020)Lin, Song, He, Wang, and Hopcroft}]{LinS00H20}
Lin, J.; Song, C.; He, K.; Wang, L.; and Hopcroft, J.~E. 2020.
\newblock Nesterov Accelerated Gradient and Scale Invariance for Adversarial Attacks.
\newblock In \emph{{ICLR}}. OpenReview.net.

\bibitem[{Liu et~al.(2017)Liu, Chen, Liu, and Song}]{Liu2016DelvingIT}
Liu, Y.; Chen, X.; Liu, C.; and Song, D. 2017.
\newblock Delving into Transferable Adversarial Examples and Black-box Attacks.
\newblock In \emph{{ICLR} (Poster)}. OpenReview.net.

\bibitem[{Liu et~al.(2022)Liu, Mao, Wu, Feichtenhofer, Darrell, and Xie}]{0003MWFDX22}
Liu, Z.; Mao, H.; Wu, C.; Feichtenhofer, C.; Darrell, T.; and Xie, S. 2022.
\newblock A ConvNet for the 2020s.
\newblock In \emph{{CVPR}}, 11966--11976. {IEEE}.

\bibitem[{Madry et~al.(2018)Madry, Makelov, Schmidt, Tsipras, and Vladu}]{MadryMSTV18}
Madry, A.; Makelov, A.; Schmidt, L.; Tsipras, D.; and Vladu, A. 2018.
\newblock Towards Deep Learning Models Resistant to Adversarial Attacks.
\newblock In \emph{{ICLR} (Poster)}. OpenReview.net.

\bibitem[{Maiya et~al.(2021)Maiya, Ehrlich, Agarwal, Lim, Goldstein, and Shrivastava}]{MaiyaFPAR21}
Maiya, S.~R.; Ehrlich, M.; Agarwal, V.; Lim, S.; Goldstein, T.; and Shrivastava, A. 2021.
\newblock A Frequency Perspective of Adversarial Robustness.
\newblock \emph{CoRR}, abs/2111.00861.

\bibitem[{Shafahi et~al.(2019)Shafahi, Najibi, Ghiasi, Xu, Dickerson, Studer, Davis, Taylor, and Goldstein}]{ShafahiNG0DSDTG19}
Shafahi, A.; Najibi, M.; Ghiasi, A.; Xu, Z.; Dickerson, J.~P.; Studer, C.; Davis, L.~S.; Taylor, G.; and Goldstein, T. 2019.
\newblock Adversarial training for free!
\newblock In \emph{NeurIPS}, 3353--3364.

\bibitem[{Sharma, Ding, and Brubaker(2019)}]{SharmaDB19}
Sharma, Y.; Ding, G.~W.; and Brubaker, M.~A. 2019.
\newblock On the Effectiveness of Low Frequency Perturbations.
\newblock In \emph{{IJCAI}}, 3389--3396. ijcai.org.

\bibitem[{Simonyan and Zisserman(2015)}]{SimonyanZ14a}
Simonyan, K.; and Zisserman, A. 2015.
\newblock Very Deep Convolutional Networks for Large-Scale Image Recognition.
\newblock In \emph{{ICLR}}.

\bibitem[{Szegedy et~al.(2017)Szegedy, Ioffe, Vanhoucke, and Alemi}]{SzegedyIVA17}
Szegedy, C.; Ioffe, S.; Vanhoucke, V.; and Alemi, A.~A. 2017.
\newblock Inception-v4, Inception-ResNet and the Impact of Residual Connections on Learning.
\newblock In \emph{{AAAI}}, 4278--4284. {AAAI} Press.

\bibitem[{Szegedy et~al.(2016)Szegedy, Vanhoucke, Ioffe, Shlens, and Wojna}]{SzegedyVISW16}
Szegedy, C.; Vanhoucke, V.; Ioffe, S.; Shlens, J.; and Wojna, Z. 2016.
\newblock Rethinking the Inception Architecture for Computer Vision.
\newblock In \emph{{CVPR}}, 2818--2826. {IEEE} Computer Society.

\bibitem[{Tram{\`{e}}r et~al.(2018)Tram{\`{e}}r, Kurakin, Papernot, Goodfellow, Boneh, and McDaniel}]{TramerKPGBM18}
Tram{\`{e}}r, F.; Kurakin, A.; Papernot, N.; Goodfellow, I.~J.; Boneh, D.; and McDaniel, P.~D. 2018.
\newblock Ensemble Adversarial Training: Attacks and Defenses.
\newblock In \emph{{ICLR} (Poster)}. OpenReview.net.

\bibitem[{Wang and He(2021)}]{Wang021}
Wang, X.; and He, K. 2021.
\newblock Enhancing the Transferability of Adversarial Attacks Through Variance Tuning.
\newblock In \emph{{CVPR}}, 1924--1933. Computer Vision Foundation / {IEEE}.

\bibitem[{Wightman(2019)}]{rw2019timm}
Wightman, R. 2019.
\newblock PyTorch Image Models.
\newblock \url{https://github.com/huggingface/pytorch-image-models}.
\newblock Accessed: 2023-12-13.

\bibitem[{Xie et~al.(2020)Xie, Tan, Gong, Wang, Yuille, and Le}]{XieTGWYL20}
Xie, C.; Tan, M.; Gong, B.; Wang, J.; Yuille, A.~L.; and Le, Q.~V. 2020.
\newblock Adversarial Examples Improve Image Recognition.
\newblock In \emph{{CVPR}}, 816--825. Computer Vision Foundation / {IEEE}.

\bibitem[{Xie et~al.(2019)Xie, Zhang, Zhou, Bai, Wang, Ren, and Yuille}]{XieZZBWRY19}
Xie, C.; Zhang, Z.; Zhou, Y.; Bai, S.; Wang, J.; Ren, Z.; and Yuille, A.~L. 2019.
\newblock Improving Transferability of Adversarial Examples With Input Diversity.
\newblock In \emph{{CVPR}}, 2730--2739. Computer Vision Foundation / {IEEE}.

\bibitem[{Xu et~al.(2019)Xu, Liu, Zhao, Chen, Zhang, Fan, Erdogmus, Wang, and Lin}]{XuLZCZFEWL19}
Xu, K.; Liu, S.; Zhao, P.; Chen, P.; Zhang, H.; Fan, Q.; Erdogmus, D.; Wang, Y.; and Lin, X. 2019.
\newblock Structured Adversarial Attack: Towards General Implementation and Better Interpretability.
\newblock In \emph{{ICLR} (Poster)}. OpenReview.net.

\bibitem[{Yao et~al.(2019)Yao, Gholami, Xu, Keutzer, and Mahoney}]{YaoGXKM19}
Yao, Z.; Gholami, A.; Xu, P.; Keutzer, K.; and Mahoney, M.~W. 2019.
\newblock Trust Region Based Adversarial Attack on Neural Networks.
\newblock In \emph{{CVPR}}, 11350--11359. Computer Vision Foundation / {IEEE}.

\bibitem[{Zhang et~al.(2023)Zhang, Tan, Sun, Zhao, Zhang, and Li}]{ZhangTSZZL23}
Zhang, Y.; Tan, Y.; Sun, H.; Zhao, Y.; Zhang, Q.; and Li, Y. 2023.
\newblock Improving the invisibility of adversarial examples with perceptually adaptive perturbation.
\newblock \emph{Inf. Sci.}, 635: 126--137.

\end{thebibliography}

% only 2-depth numbered titles for Appendix
\appendix
\setcounter{secnumdepth}{2}
\section{Appendix}

\subsection{The Impact of Chroma Channels}

We primarily evaluated the impact of luma channel's $r_Y$ over adversarial transferability in our paper. Our assumption is that the luma channel holds more structural information than that of the chroma channels, thereby being the more valuable channel for centralizing perturbation. This is also our rationale for choosing the quantization ratios of $r_Y=0.9, r_{Cb}=0.05, r_{Cr}=0.05$ as defaults. Experimental results prove that centralizing more perturbation towards the luma channel yields greater adversarial effectiveness.

In this section, we further investigate the effect of individual chroma channels, namely the Cb channel and the Cr channel. Following our previous evaluation settings, where the cumulative quantization rate of $1/3$ remains constant, we iterate $r_{Cb}$ and $r_{Cr}$ each from $0$ to $1$ with a step size of $0.1$, allocating the rest equally to the remaining two channels respectively.

\begin{figure}[htbp]
    \centering
    \includegraphics[width=\linewidth]{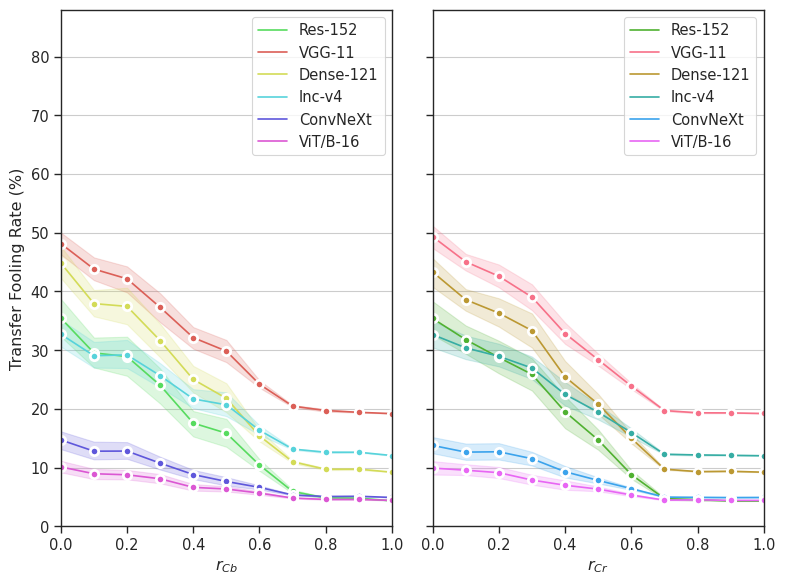}
    \caption{The effect of the quantization ratio of chroma channels $r_{Cb}$ (left) and $r_{Cr}$ (right) over adversarial transferability respectively.}\label{fig:effect-of-r-chroma}
\end{figure}

Figure~\ref{fig:effect-of-r-chroma} illustrates that neither increasing $r_{Cb}$ nor $r_{Cr}$ improves adversarial transferability. Fooling rates decrease at a similar rate as quantization ratios for channel Cb and Cr increases respectively. Adversarial transferability never manages to reach our default $r$ settings, falling less than 50\% at best.

Our findings indicate:
\begin{enumerate}
    \item There is no significant difference in the impact of the chroma channels, i.e., channel Cb and Cr, on adversarial effectiveness.
    \item Centralizing perturbation towards the chroma channels results in a deteriorated adversarial transferability, consistently.
\end{enumerate}

As such, our default strategy of preserving the luma channel to the largest extent ($r_Y=0.9$) and allocating the remaining quantization budget evenly to the two chroma channels ($r_{Cb} = 0.05, r_{Cr} = 0.05$) is a rational approach.

\subsection{Details of Ablation Study}

In our ablation study, we evaluated our approach with 4 other strategies, namely: \textit{RandA}, \textit{RandB}, \textit{Low}, and \textit{High}. We address the details of these frequency quantization strategies by illustrating the quantization matrices $Q_Y, Q_{Cb}, Q_{Cr}$ of these schemes that we consider.

To begin, we acknowledge the \textit{zig-zag} sequence, which, for $(8\times 8)$ frequency blocks, is formulated as
\begin{equation}
    \label{eq:zigzag}
    \begin{bmatrix}
        0  & 1  & 5  & 6  & 14 & 15 & 27 & 28 \\
        2  & 4  & 7  & 13 & 16 & 26 & 29 & 42 \\
        3  & 8  & 12 & 17 & 25 & 30 & 41 & 43 \\
        9  & 11 & 18 & 24 & 31 & 40 & 44 & 53 \\
        10 & 19 & 23 & 32 & 39 & 45 & 52 & 54 \\
        20 & 22 & 33 & 38 & 46 & 51 & 55 & 60 \\
        21 & 34 & 37 & 47 & 50 & 56 & 59 & 61 \\
        35 & 36 & 48 & 49 & 57 & 58 & 62 & 63
    \end{bmatrix}
\end{equation}
This is the standard for JPEG encoding, and is the sequence for frequency coefficient blocks from low-frequency to high-frequency.

\begin{figure}[htbp]
    \centering
    \includegraphics[width=\linewidth]{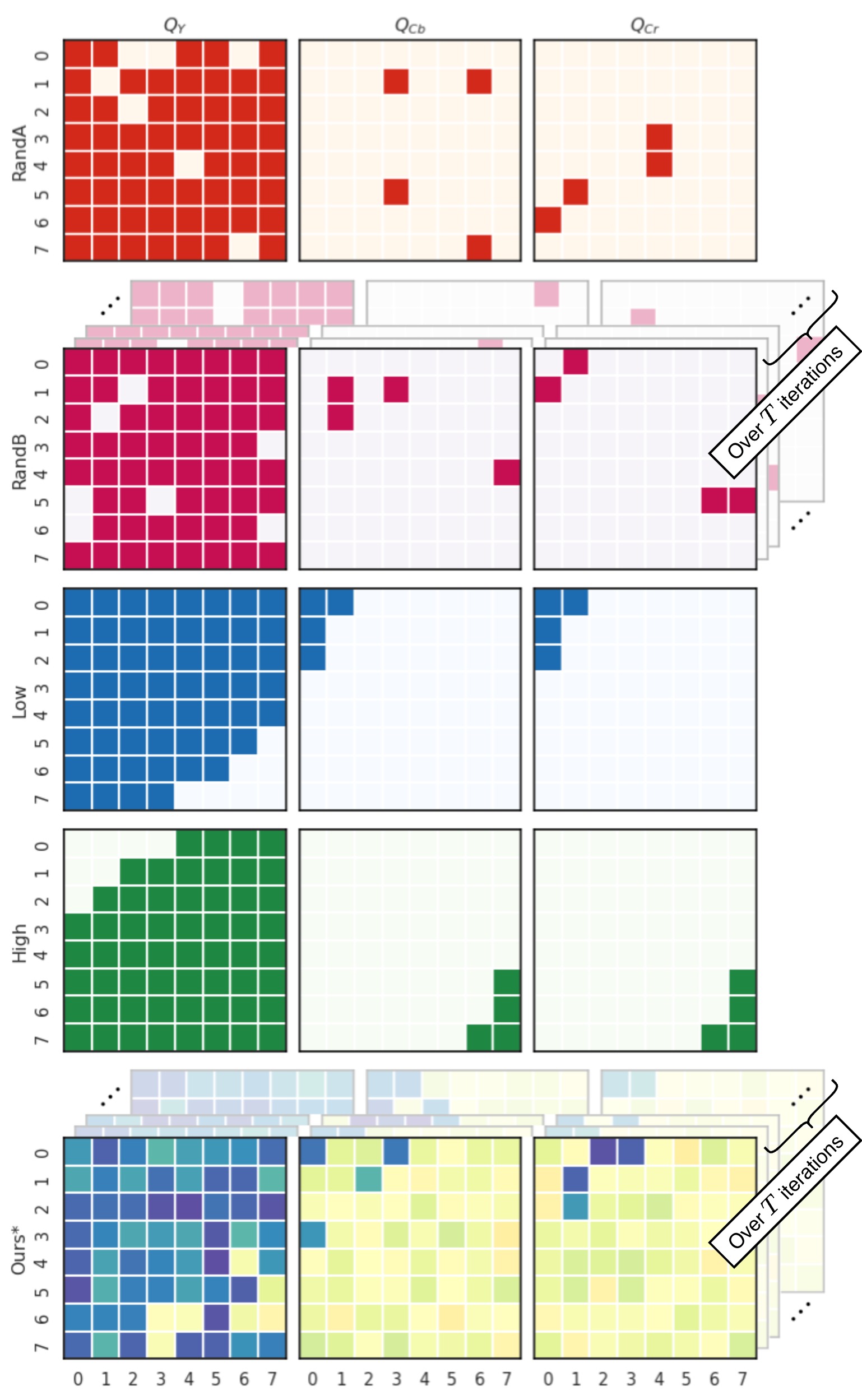}
    \caption{The 4 strategies (top 4 rows) that we evaluate against our method (bottom row) in the ablation study. Note that \textit{RandB} and \textit{Ours} consists of a sequence of matrices, which we use to illustrate that the matrices change dynamically over iterations. In contrast, \textit{RandA}, \textit{Low}, and \textit{High} all use fixed matrices.}\label{fig:ablation-strats}
\end{figure}

In Figure~\ref{fig:ablation-strats}, we illustrate quantization schemes evaluated in our ablation study. We plot the heatmap of demonstrations of the quantization matrices used in each strategy. The darker colors represent $1$, i.e., frequency coefficient blocks that are preserved, whereas the lighter colors represent $0$, i.e., blocks that are eliminated.

\begin{itemize}
    \item In strategy \textit{RandA} and \textit{RandB} (top 2 rows), blocks are eliminated at random.
    \item In strategy \textit{Low} and \textit{High} (row 3 and 4), lower and higher frequency coefficient blocks are preserved respectively, with respect to the sequence defined in Matrix~\ref{eq:zigzag}.
    \item Lastly, in our strategy (bottom row), we are able to acquire the importance of each frequency coefficient block (illustrated as different shades of colors) at each attack iteration through back-propagation, thereby implementing our dynamic and guided fine-grained centralization of adversarial perturbation.
\end{itemize}

We layer multiple matrices over each other in strategy \textit{RandB} and \textit{Ours} in Figure~\ref{fig:ablation-strats}, so as to represent the dynamicality of the process, where quantization differs at each iteration. Where the rest of the strategies, namely \textit{RandA}, \textit{Low}, and \textit{High} all use fixed quantization matrices defined at the start of the attack.

The results presented in our ablation study demonstrate that the lower frequency coefficient blocks exert greater dominance, proving that concentrating perturbations in these areas would enhance adversarial effectiveness.

Interestingly, we discovered that solely preserving the higher frequency regions leads to a catastrophic failure of the attack, resulting in fooling rates dropping to zero in most cases.

Finally, as depicted, our strategy surpasses the brute centralization approach of \textit{Low} by allowing for dynamically adjustments to quantization, facilitating a better alignment with model predictions.

\subsection{Details on Differential Quantization}

\begin{listing}[htbp]
    \centering
    \caption{Differential Implementation of $\mathcal{R}(\cdot)$}%
    \label{lst:listing}%
    \begin{lstlisting}[language=Python]
def quantize(x: torch.Tensor, ratio: float = 0.9) -> torch.Tensor:
  # Calculate the threshold value of the top `ratio`
  threshold = torch.quantile(x, 1 - ratio)

  # Keep all values above the threshold as 1
  x_hat = torch.zeros_like(x)
  x_hat[x >= threshold] = 1

  # A hack to make the function differentiable
  return x_hat - x.detach() + x
\end{lstlisting}
\end{listing}

Equation~\ref{eq:6} in our paper presents us applying rounding $\mathcal{R}(\cdot)$ and acquiring the quantization matrix in a non-differential manner, as the function is a staircase, hindering back-propagation based optimization. To solve this issue, we implemented the function as in Listing~\ref{lst:listing}.

\end{document}